# Telemedicine as a Special Case of Machine Translation

Krzysztof Wołk [1], Krzysztof Marasek [1] and Wojciech Glinkowski [2,3]

[1] Department of Multimedia

Polish - Japanese Academy of Information Technology

[2] Chair and Department of Orthopedics and Traumatology of the Locomotor System, Center of Excellence "TeleOrto" for Telediagnostics and Treatment of Disorders and Injuries of the Locomotor System, Medical University of Warsaw

[3] Polish Telemedicine Society, Warsaw, Poland

**Corresponding Author:**

Wojciech Glinkowski MD, Ph.D.

Chair and Department of Orthopedics and Traumatology of the Locomotor System, Center of Excellence "TeleOrto" for Telediagnostics and Treatment of Disorders and Injuries of the Locomotor System, Medical University of Warsaw

Lindleya str.4

02-005 Warszawa

POLAND

Tel: +48225021197

Fax: +48225022100

E-mail: w.glinkowski@gmail.com

**Introduction**

Statistical Machine Translation (SMT) is the translation of text by a computer, with no human involvement. SMT systems have no knowledge of language rules. Instead, they "learn" to translate by analyzing large amounts of data for each language pair. They can be trained in specific industries or disciplines using additional data relevant to the sector needed. Typically, SMT systems deliver fluent-sounding but less consistent translations.

The machine translation is evolving quite rapidly in terms of quality. Currently, several machine translation systems are available on the web that provide reasonable translations. Developed systems are not perfect, and their quality may decrease in some specific domains. In addition to this, the scientific community is involved in machine translation. It must be pointed out that the scientific organizations, conferences, and events dedicate great effort to its improvement. One of the biggest advantages of machine translation is that most users do not require perfect translations [1]. Users may only be interested in roughly understanding a text simply to get an idea of what the text is about. However, other users may not be that flexible. For example, the correctness and beauty of writing in medicine may not be important, but the precision and adequacy in the translated message is crucial. In medical communication, a translation error between the patient and the physician, or an error in communication regarding treatment or a diagnosis may have serious consequences for a patient's health [2].

Recently, due to the growing success of and interest in language technologies, machine translation has been applied to the field of medicine. For example, one study [3] analyzed the feasibility of post-editing machine translations of health-promotional English documents from local and national public health websites in the USA. It was assumed, a priori, that machine translation would not provide a high enough quality for the documents to be used as official versions. Despite that, language technologies are steadily increasing in quality. It should be expected that, in the not-too-distant future, machine translation will be capable of translating any text in any domain with the required quality.

The medical data domain is, in our opinion, a very narrow, but relevant and promising field of research for language technologies. MT systems can be used for translation of medical records of any kind. Accessing and translating a foreign patient's medical history might even save their life. Preparation of direct speech-to-speech translation systems is also possible. The foreign patient's speech is recognized using an Automated Speech Recognition (ASR) system. After recognition, the speech is translated into another language and synthesized in real-time. For example, the EU-BRIDGE project aims at developing automatic transcription and translation technology that will permit the development of innovative multimedia captioning and translation services of audiovisual documents between European and non-European languages [http://www.eu-bridge.eu].

Obtaining and providing medical information in comprehensive ways appears to be of crucial importance for both patients and physicians [4-7]. For example, as emphasized by Healthcare Technologies for the World Traveler (HTH) [8], a foreign patient may require an explanation and description of their diagnosis and comprehensive information about available treatment options. In several countries, many residents and immigrants communicate in languages other than the official one.

According to Karliner et al. [9], it is necessary to analyze how human translators could enhance access to health care, including improvement of its quality [10]. Nevertheless, human translators experienced in telemedicine information are very often unavailable for both patients and medical professionals [11]. Although existing machine translation capacities are imperfect [11], machine translation must ensure the reduction of costs associated with medical translation. On the other hand, it is necessary to increase its availability and quality [12].

Medical professionals, researchers, and patients require adequate access to the abundance of telemedicine information on the Internet [6, 13]. This information can potentially improve our health and well-being. Sharing medical information could improve medical research, as well. English is the most dominant language used in medical science, but not the only one.

Polish is considered to be the one of the most challenging West-Slavic languages, due to its complexity. It is a tough language for an SMT system. Polish grammar, for example, includes complicated rules and elements, including an immense vocabulary (thanks to its complex declension). Nearly free word order in sentences is also problematic. All of these are the main reasons for its challenging character. In addition, the Polish language includes 7 cases and 15 gender forms for both nouns and adjectives.

As expected, these facts strongly influence the data and data structure used in statistical translation models. The lack of available and appropriate resources necessary for data input to SMT systems presents another problem. SMT systems give the best results for concrete and narrow text domains. The proper quality of the parallel data, including the required domains, has inadequate availability. On the other hand, Polish and English differ strongly in syntax. Above all, English is a positional language. This means that the syntactic order, which includes the word order of one sentence, has an invaluable significance, especially because of the limited inflection of words (for example, lack of declension endings). Sometimes, the position of the word in a sentence is the only indicator of the sentence meaning. As far as English sentences are concerned the subject comes before the predicate. Therefore, a sentence is structured in Subject-Verb-Object (SVO) word order. In contrast, Polish simply has no particular word order. Additionally, the word order itself has no decisive impact on a sentence's meaning. In Polish, one can express the same idea in many ways, which is simply not possible in English. For instance, the sentence "I have bought myself a new car." can be expressed in Polish as "Kupiłem sobie nowy samochód", or "Nowy samochód sobie kupiłem.", or "Sobie kupiłem nowy samochód.", or "Samochód nowy sobie kupiłem." As one can see, changes in word order influence the complexity of the translation process.

As a consequence, the development of SMT systems for the Polish language has been considerably slower in comparison to English and other languages. The primary goal of this research is to develop an SMT system for translation from Polish to English language and vice versa, with an emphasis on medical data. This paper has the following structure: Section 2 contains an introduction to the preparation of Polish data. Section 3 presents the English language issues. Section 4 describes the methods associated with translation evaluation. Section 5 presents the results. Sections 6 and 7 provide the summary of potential implications and opportunities for future work.

**Polish Data Preparation**

The Polish data we included was a corpora derived from the European Medicines Agency (EMEA) parallel corpus. This corpus was created from biomedical PDF documents from the agency. It includes documents related to medical products and their translations into 22 official languages of the European Union. It contains roughly 1,500 documents for most of the languages, but not all of them are available in every language [14]. It comprises around 80 MB of data and 1,044,764 sentences constructed from 11.67M words that were not tokenized. The data is pure text encoded in UTF-8. Additionally, the texts were separated into sentences (one per line) and structured in language pairs. We chose this corpus as most similar to medical texts (which we did not have access to in sufficient quantity) in terms of complexity and vocabulary.

The vocabulary consisted of 148,170 unique Polish and 109,326 unique English tokens [15]. The disproportionate vocabulary size and number of tokens are also a true challenge, especially when it comes to translation from English to the Polish language.

Before the use of the training translation model, preprocessing that included removal of long sentences (set to 80 tokens) had to be performed to limit the model size and computation time. Moses toolkit scripts [16] were used for this purpose. Moses is an open-source toolkit for statistical machine translation that supports linguistically-motivated factors, confusion network decoding, and efficient data formats for translation models and language models. In addition to the SMT decoder, the toolkit also includes a wide variety of tools for training, tuning, and applying the system to many translation tasks.

**English Data Preparation**

The preparation of the English data was far less complicated than that of the Polish data. We developed a tool to clean the English data by eliminating foreign words, strange symbols, etc. Compared to Polish, the English data included drastically fewer errors. However, some problems needed to be fixed. The most problematic were translations into languages other than English itself, including strange UTF-8 symbols, repetitions, and unfinished sentences. Such errors are typical when corpora are built using automatic tools.

**Methods of Evaluation**

Human evaluations of machine translation outputs require considerable effort and are expensive. Human evaluations can take days or even weeks to finish. So, automatic metrics are needed to measure the translation quality derived from SMT systems. Different automated metrics are used to compare SMT translations and match the human translations. Among the most widely used SMT metrics are:
- the Bilingual Evaluation Understudy (BLEU),
- the U.S. National Institute of Standards & Technology (NIST) metric;
- The Metric for Evaluation of Translation with Explicit Ordering (METEOR), and
- Translation Error Rate (TER).

SMT metrics were briefly described by Radziszewski [17]. BLEU is quick to use, is inexpensive to operate, is language independent, and correlates highly with human evaluation. It is the most widely-used automated method of determining the quality of machine translation. The metric scores of translation range from 0 to 1. It is frequently displayed as a percentage value. The closer to 1 (100%), the more the translation correlates to human translation. Put simply, the BLEU metric measures how many words overlap in a given translation when compared to a reference translation, giving higher scores to sequential words. Scores lower than 15% mean that the machine translation engine is unable to provide satisfactory quality, as reported by Lavie [19] and a commercial software manufacturer [18]. A high level of post-editing will be required to finalize output translations and reach publishable quality.

Fig.1 BLEU interpretability scale.

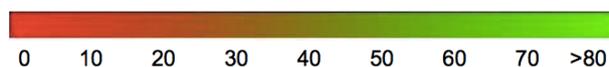

A system score greater than 30% means that translations should be understandable without problems. Scores over 50 reflect good and fluent translations.

The general approach for BLEU, as described in [20], is to attempt to match variable length phrases to reference translations. Weighted averages of the matches are then used to calculate the metric. The use of different weighting schemes leads to a family of BLEU metrics, such as standard BLEU, Multi-BLEU, and BLEU-C [21].

The basic BLEU metric is expressed by equation [7]:

$$BLEU = P_B \exp\left(\sum_{n=0}^{N} w_n \log p_n\right)$$

where $p_n$ is an *n*-gram precision using *n*-grams up to length N and positive weights $w_n$ that sum to one. The brevity penalty $P_B$ is calculated as:

$$P_B = \begin{cases} 1, & c > r \\ e^{(1-\frac{r}{c})}, & c \leq r \end{cases}$$

where *c* is the length of a candidate translation, and *r* is the effective reference corpus length [20].

The standard BLEU metric calculates the matches between *n*-grams of the SMT and human translations, without considering position of the words or phrases within the texts. In addition, the total count of each candidate SMT word is limited by the corresponding word count in each human reference translation. The BLEU metric avoids bias that would enable SMT systems to overuse high-confidence words to boost their score. BLEU applies this approach to texts sentence-by-sentence, and then computes a score for the overall SMT output text. In doing this, the geometric mean of the individual scores is used, along with a penalty for excessive brevity in translation [20].

On the other hand, the NIST metric tends to improve the BLEU metric by evaluating information in several ways. It uses the arithmetic mean vice the geometric mean of the *n*-gram matches to emphasize the proper translation of rare words. The NIST metric also puts stronger importance on rare words. This metric has shown improvements compared to the BLEU metric. The NIST metric can output values between 0 and 15; the higher value, the better translation quality [22].

The METEOR metric, introduced by the Language Technologies Institute of Carnegie Mellon University, is also designed to improve the BLEU metric. We applied it without synonyms and paraphrased matches for Polish. METEOR emphasizes recall by changing the BLEU brevity penalty. In addition, it takes into consideration higher order *n*-grams to favor matches in word order, and uses arithmetic vice geometric average values, as well. For multiple reference translations, METEOR measures the best score for word-to-word matching. Similar to BLEU, this metric also returns scores between 0 and 100. A detailed description of this metric was provided by Doddington [23].

TER is the most recent SMT metric developed. This metric evaluates the minimum number of human corrections necessary for an SMT translation to successfully match a reference translation in terms of meaning and fluency. Necessary human corrections might include insertion, removal, and changing of words or phrases. In contrast to other metrics, a translation with a lower TER value is more similar to the reference translation. TER scores range from 0 to 100 [24].

A number of experiments have been conducted to evaluate different versions of SMT systems. The experiments included several steps, including corpora processing, tokenization, cleaning, factorization, lower casing, splitting, and final cleaning. Training data were evaluated, and a language model was developed. Tuning was performed for every experiment. Finally, the experiments were performed.

Testing was performed using the Moses open source SMT toolkit, including the Experiment Management System (EMS) [25]. The SRI Language Modeling Toolkit (SRILM) [26], along with an interpolated version of Kneser-Ney discounting (interpolate –unk –in discount), was used for 5-gram language model training. We used the MGIZA++ tool for word and phrase alignment. KenLM [27] was applied to ensure the binarization of the language model. Lexical reordering was set to mid-bidirectional-fee. Reordering of the phrases' probabilities was performed according to the lexical values of the phrases. It includes three different types of orientation, with an emphasis on sources and targeted phrases: monotone (M), swap(S), and discontinuous (D). The reordering of the bidirectional models includes the possible probabilities of the mutual positions of sourced counterparts in correlation with the actual and subsequent phrases. The probability distribution of foreign phrases is evaluated by "f," and English phrases by "e" [2, 28-30]. MGIZA++ is a multi-threaded version of the famous GIZA++ tool [26].

A method of symmetrizing was developed to ensure appropriate word alignment. First, two-way alignments derived from GIZA++ were structured. As a result, only the points of alignments that appeared in both alignments were left. In the next phase, additional points of alignments that appear in their union were combined. The other steps contribute to the potential alignment points of unaligned and neighboring words. Neighboring can be positioned directly to left or right, top or bottom, including a diagonal (grow-dialog) position. During the final phase, the points of alignments between the words are combined (grow-dialog-final), where at least one is unaligned. If the grow-dialog-final method is applied, the point of alignment between the two unaligned words eventually occurs [31].

Descriptive statistics, parametric, non-parametric tests, and inter-rater correlations were performed using the MedCalc Statistical Software version 15.2 (MedCalc Software bvba, Ostend, Belgium; http://www.medcalc.org; 2015) to find differences between the applied scores and between translation directions.

**Results**

We executed the following experiments to evaluate the optimal translation method from Polish to English, and vice versa. These experiments were performed with test and development data. Data was obtained through the process of random selection and removal from the corpora itself. We have cumulated 1000 sentences for each case. The BLEU, NIST, TER, and METEOR metrics evaluated the results of these experiments. It is worth mentioning that a low value of the TER metric tool is considered to be a better one, while the other metrics indicate higher quality when their values are higher. All metrics were normalized to fit in the range of 0 to 100 (the higher value, the better) for easier comparison and understanding. Figures 2 and 3 depict the results for each experiment. The metrics were averaged for a more straightforward interpretation. Tables 1 and 2 illustrate the experimental results. Experiment 00 in these tables signifies the baseline system. Every test involved a separate modification of the baseline. In each experiment, we seek improvements in comparison with the baseline system. In addition, Experiment 01 relies on true casing and punctuation normalization.

Experiment 02 is improved with the help of the Operation Sequence Model (OSM). The reason for introducing the OSM is the provision of phrase-based SMT models, which can memorize dependencies and lexical triggers. In addition, the OSM uses a source and target context, with the exclusion of the spurious phrasal segmentation problems. The OSM is invaluable, especially for the strong mechanisms of reordering. It combines both translation and reordering, handles short and long-distance reordering, and does not need a reordering limit [32].

Experiment 03 includes a factored model that provides an additional annotation of the word levels, which may be exploited in different models. We evaluate the part of speech tagged data in correlation with the English language segment as a basis for the factored phrase models [33].

Hierarchically-structured phrase-based translations leverage the strengths of both phrase-based and syntax-based translations. They use phrases (word segments or blocks) as translation units, including synchronous context-free grammar cases as rules (syntax-based translations). Hierarchically-structured phrase models enforce the rules with gaps. Since these are illustrated by non-terminals and the rules themselves are best evaluated with a search algorithm, this is similar to syntactic chart parsing. These models can be categorized as a class of tree-based or grammar-based models. We applied such a model in Experiment 04.

The Target Syntax model includes the application of linguistic annotation for non-terminals in the hierarchically-structured models. The application requires a syntactic parser. In this case, we applied the "Collins" [34] statistical parser of natural language in Experiment 05.

Experiment 06 was executed with the stemmed word alignment. The factored translation model enabled it to determine word alignment based on word structure that differs from surface word formations. One very popular method is to apply stemmed words to these alignments of words. There are two main reasons for such an approach. For morphologically-rich languages, stemming deals with the data parity problem. On the other hand, GIZA++ may face serious challenges with the immense vocabulary, since stemming influences the number of unique words.

Experiment 07 applies Dyer's Fast Align [35], which is an option in GIZA++. It works much faster with the better results, especially for language pairs, which come with no need for the large-scale reordering.

In Experiment 08 we applied settings recommended by Koehn for his SMT system at WMT'13 [36]. In Experiment 09, we changed the language model discounting to Witten-Bell [37]. This discounting method considers the diversity of predicted words. It was developed for text compression and can be considered an instance of Jelinek-Mercer smoothing. The nth order smoothed model is defined recursively as a linear interpolation between the nth order maximum likelihood model and the (n-1)th order smooth model [37].

Lexical reordering was set to her-user-bidirectional-be in Experiment 10. This is a hierarchical reordering method that considers different orientations: monotone, swap, discontinuous-left, and discontinuous-right. The reordering is modeled bidirectionally, based on the previous or next phrase, conditioned on both the source and target languages.

Compounding is a method of word formation consisting of a combination of two (or more) autonomous lexical elements that form a unit of meaning. This phenomenon is common in German, Dutch, Greek, Swedish, Danish, Finnish, and many other languages. For example, the word "flowerpot" is a closed or open compound in English texts. It results in a lot of unknown words in any text, so splitting up these compounds is a standard method when translating from such languages. Moses offers a support tool that splits up words if the geometric average of the frequency of its parts is higher than the frequency of a word. In Experiment 11, we used the compound splitting feature. Lastly, in Experiment 12, we used the same settings as for the out-of-domain corpora in IWSLT'13 [38].

In Experiment 13, we extended the parallel corpora with dialogs from movies obtained from the OPUS project. We found this necessary, because the EMEA corpora lacked casual speech data, which would most likely prove necessary for communication between humans. Moreover, it was supposed to increase the quality of in-domain data translation. The additional corpora consisted of 23,825,341 sentence pairs built from over 117M words. We applied that corpus to Experiment 4, because it provided the best translations in other experiments [39].

Lastly, we performed Experiment 14. English-to-Polish translation is usually harder and has poorer quality. We interpolated the Google WEB1T language model [40] into our system using the SRILM tool.

**Tab.1 Polish to English translations results**

| System | BLEU | NIST | METEOR | TER |
|---|---|---|---|---|
| 00 | 70.15 | 70.23 | 82.19 | 70.62 |
| 01 | 64.58 | 65.16 | 76.04 | 64.38 |
| 02 | 71.04 | 70.76 | 82.54 | 71.67 |
| 03 | 71.22 | 70.56 | 82.39 | 71.49 |
| 04 | 76.34 | 73.30 | 85.17 | 75.23 |
| 05 | 70.33 | 70.36 | 82.28 | 70.73 |
| 06 | 71.43 | 70.70 | 82.89 | 71.27 |
| 07 | 71.91 | 71.76 | 83.63 | 73.40 |
| 08 | 71.12 | 69.16 | 84.55 | 70.05 |
| 09 | 71.32 | 71.36 | 83.31 | 72.32 |
| 10 | 71.35 | 69.36 | 81.52 | 70.26 |
| 11 | 70.34 | 70.96 | 82.65 | 71.78 |
| 12 | 72.51 | 71.36 | 82.81 | 71.81 |
| 13 | 76.97 | 73.83 | 86.11 | 76.40 |

**Tab.2 English to Polish translation results**

| System | BLEU | NIST | METEOR | TER |
|---|---|---|---|---|
| 00 | 69.18 | 67.63 | 79.21 | 69.61 |
| 01 | 61.15 | 61.29 | 71.91 | 60.55 |
| 02 | 69.41 | 67.63 | 78.98 | 69.10 |
| 03 | 68.45 | 67.10 | 78.63 | 68.38 |
| 04 | 73.32 | 69.90 | 81.72 | 72.95 |
| 05 | 69.21 | 67.70 | 79.26 | 69.12 |
| 06 | 69.27 | 67.76 | 79.30 | 68.73 |
| 07 | 68.43 | 67.16 | 78.95 | 66.95 |
| 08 | 67.61 | 65.83 | 77.82 | 70.05 |
| 09 | 68.98 | 67.43 | 78.90 | 68.87 |
| 10 | 68.67 | 66.83 | 78.55 | 68.08 |
| 11 | 69.01 | 67.63 | 79.13 | 69.16 |
| 12 | 67.47 | 65.96 | 77.65 | 66.68 |
| 13 | 73.67 | 71.83 | 82.13 | 73.17 |
| 14 | 74.10 | 72.16 | 82.43 | 74.03 |

Fig.2 Polish to English Experiments results.

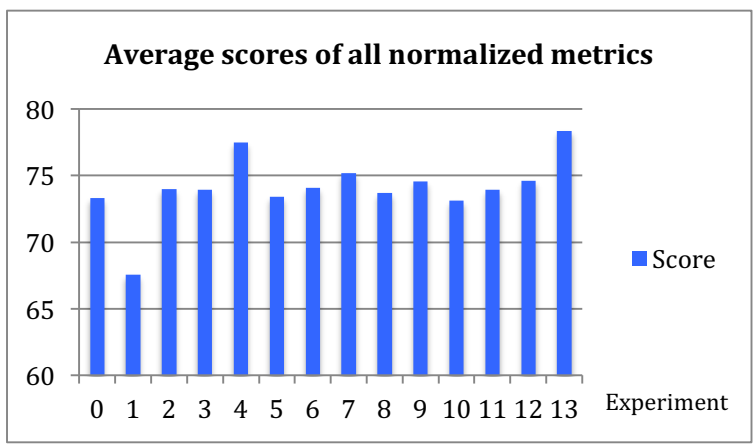

Fig.3 English to Polish Experiments results.

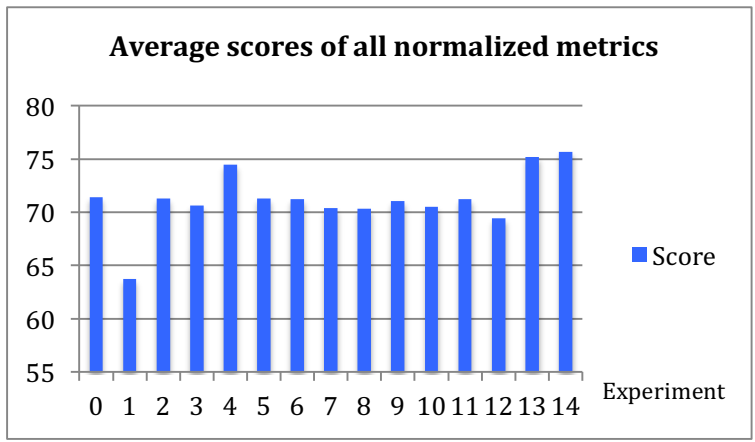

Normalized metrics were used to compare results. Scores lower than 15% mean that the machine translation engine was unable to provide satisfactory quality. Scores greater than 30% mean that translations should be understandable without problems. Scores over 50% reflect adequate translations. The average results of Polish-to-English translations scores for BLEU, NIST, METEOR, and TER were relatively high, ranging from 70.58 to 82.72. The lowest score was 64.38. The average results for English-to-Polish translations were a little lower (67.58-78.97). The Wilcoxon Matched Pairs Test revealed statistically significant differences between BLEU and METEOR scores and BLEU and NIST scores (at $p < 0.05$). No significant differences were observed between BLEU and TER scores for Polish-to-English translations. The Wilcoxon Matched Pairs Test revealed statistically significant differences between BLEU and NIST scores and BLEU and METEOR scores. No significant differences were found for BLEU and TER scores for English-to-Polish translations.

The model concerning the same raters for all subjects in two-way model tested for absolute agreement revealed that measurements of Intraclass Correlation Coefficient for BLEU scores for translations in both directions ranged from 0.6531 to 0,9149 for single measurements and from 0.7901 to 0.9556 for average measurements.

The measured Intraclass Correlation Coefficient of METEOR scores ranged from 0.3631 to 0.7783 for single measurements and from 0.5328 to 0.8754 for average measures. The same method for NIST scores presented 0.4162 to 0.8140 for single measurements and from 0.5878 to 0.8975 for average measurements. Finally, the measurements of the Intraclass Correlation Coefficient for TER scores ranged from 0.5666 to 0.8750 for single measurements and from 0.7233 to 0.9333 for average measurements. The Student t-test revealed significant differences between TER scores calculated for English–Polish translations vs. Polish–English translations. The difference was 2.5 (S.E. 1.2085) at $p = 0.0324$.

Such differences are not surprising because of the differences in all metrics estimation methods. Especially the METEOR additionally scores words that are synonymic in their meaning. The NIST alters BLEU scores by additional calculation of how informative particular n-grams were. The statistical significance between PL to EN and EN to PL translation was found and can be considered normal because of the greater complexity and bigger vocabulary of the Polish language.

**Discussion and Conclusions**

A couple of conclusions can be directly drawn from the results of the experiments conducted in this research. It was a little bit surprising that true casing and punctuation normalization lowered the scores by a significant factor. We believe that the texts have already been adequately cased and punctuated. In Experiment 02, we noticed that OSM lowered some metrics results. It often increases the quality of translations. Nevertheless, in the PL->EN translation experiment the BLEU score improved just slightly, but, on the other hand, other metrics decreased. Similar results can be seen in the EN->PL translation experiments. In this case, the BLEU score improved, but the other metrics decreased.

The majority of other experiments produced the expected results. Almost all of them improved the score a little bit or at least confirmed our conclusions with each metric. Unfortunately, Experiment 12, which was structured in settings that ensured the best system score in IWSLT 2013 evaluation, did not improve the quality of this data as much as it previously did. The most probable reason is that the data applied in IWSLT did not derive from any specific text domain, while here we had to deal with a very narrow domain. The training, tuning, and adjustment parameters may need to be adjusted separately for every text domain. As for the cases in which improvements cannot be replicated, this should be done first. On the other hand, improvements achieved by training the hierarchically-based model surprised us. Compared to other experiments, Experiment 04 improved the BLEU score significantly. In addition, significant improvements can be noticed in both the PL->EN and EN->PL language translations, which most likely provide an excellent starting point for future research and experiments.

It was decided to use that system for Experiments 13 and 14 with additional parallel data and Google's language model for Polish. The anticipated improvement of translation quality was achieved in both of these experiments. The translation quality in Experiment 13 is very close to what was anticipated, based on out-of-domain data. It was required to cover any out-of-domain translation scenarios or casual human dialogs. What is worth mentioning here is that, in our final experiments, we were able to obtain almost equal quality of translation in both directions.

Translation from EN to PL was much harder, which was proved by the experiments. As expected, the results of translation into Polish were not that good. Very likely, the reason for this is the complex structure of Polish grammar including the vast Polish vocabulary. It was one of the reasons why we used the Google language model.

Our analysis led us to think that the translation results, where the BLEU measure is greater than 70, can be accepted as satisfactory only within the limits of a selected text domain. Such results should be possible to replicate with any language pair, but especially West-Slavic because of the linguistic and lexical similarities. Considering the specification of the BLEU metric, we assume that our translation systems perform with truly high quality, but there is still room for improvement. The differences of translation output (T) vs. original (O) texts are presented in Table 3. We can see that mistakes preserve sentence meaning and simply use synonyms or different sentence structure in many cases. The evaluation scores indicate that people could adequately comprehend the translations and that they are good enough to help them in their work, but they are not sufficient to be used in such specialized areas like medicine yet (for example, in the case of patients in foreign hospitals). We have every reason to believe that an improvement of the BLEU score to a threshold higher than 80, or perhaps 85, could create systems that could be used for practical cases. It may be especially useful with the Polish language, which is one of the most complex in terms of its structure, grammar, and spelling. Additionally, it must be emphasized that the experiments were conducted on texts obtained from the PDF documents shared by the European Medicines Agency. That is the reason the data is more complex and uses a more sophisticated vocabulary than casual human speech. Speech is typically less complicated and easier to process by SMT systems. It is here that we see another opportunity to increase output translation quality. Comparable results could be obtained for other language pairs as long as text domain is limited. Nonetheless, it would require adaptation of data and training parameter for each language independently.

**Tab.3 SMT Output**
Table presents translation (T) output (the result of machine translation) and the original reference sentence (O) texts. Differences are underlined. Sentences were randomly selected by an automatic algorithm and equally from each batch of a text during the evaluation of machine translation. The selection was not performed manually, to avoid bias.

| Sentence | Text |
|---|---|
| 1 | **O:** I think we should reflect carefully before we do that again.<br><br>**T:** I believe that we should reflect carefully before we do that again. |
| 2 | **O:** International cooperation is an essential tool for an effective fight against this scourge, at both national and multilateral levels.<br><br>**T:** International cooperation is an essential tool for an effective fight against this scourge, at both national and international levels. |
| 3 | **O:** We have completely overhauled our policy style and the way in which we prepare and review legislation.<br><br>**T:** We have completely overhauled our policy style and the way in which we prepare and review legislation. |
| 4 | **O:** We have new objectives for various trades here.<br><br>**T:** We have new objectives for various trades here. |
| 5 | **O:** As regards subheading 1a: Competitiveness for growth and employment the Council intends to provide for appropriate funding of priorities connected with the Lisbon Strategy, having left a significant margin to cover the European Union priorities defined in the conclusions of the June European Council, as well as other so far unforeseen needs.<br><br>**T:** As regards sub-heading 1a: Competitiveness for growth and employment, the Council intends to provide for appropriate funding of priorities connected with the Lisbon Strategy, left a significant margin to cover the European Union priorities defined in the conclusions of the June European Council, as well as others so far unforeseen needs. |
| 6 | **O:** As far as I am aware, a rather successful long-term social dialog has been conducted on these issues and in my opinion, it is thus necessary to reconcile seemingly contradictory procedures.<br><br>**T:** As far as I know, in this regard have been carried out rather successful long-term social dialog and, in my view, there is a need to reconcile |
| | seemingly contradictory procedures. |
| 7 | **O:** They are concerned with the creation of a milk fund that will support reforming activities in this sector, such as the promotion of milk consumption, including consumption in schools, or support for milk production in <u>mountainous</u> areas.<br><br>**T:** They are concerned with the creation of a milk fund that will support reforming activities in this sector, such as the promotion of milk consumption, including consumption in schools, or support for milk production in <u>mountain</u> areas. |

**Future Work**

Using machine translation for medical texts has a high potential for providing benefits to patients, including tourists and people who do not know the language of the country in which they require medical help. Improved access to various medical information can be very profitable for patients, medical professionals, and eventually to medical researchers.

Human interpreters with proper medical training are extremely rare and costly. Machine translation could also assist in the evaluation of medical history, diagnoses, adequate medical treatments, health-related information, and the findings of medical researchers around the world. Mobile devices, the Internet, and web applications can be used to boost the delivery of machine translation services for medical purposes, even as real-time speech-to-speech services.

Several potential opportunities for future work are of interest, helping us to extend our research in this critical area. Additional experiments using extended language models are warranted to improve the SMT scores. Much of the literature [27] confirms that interpolation of out-of-domain language models and adaptation of additional bilingual corpora would improve translation quality. We intend to use linear interpolation, as well as Modified Moore Levis filtering for these purposes. We are also interested in developing some web crawlers to obtain additional data that would most likely prove useful. Good quality, parallel data, especially in the required domain, has low availability.

In English sentences, the subject group precedes the predicate, so the sentences are ordered according to Subject-Verb-Object (SVO) word order. Modification of the Polish data to SVO order can comprise an interesting experiment in the future, as well.

**Acknowledgement**

This work was supported by the European Community from the European Social Fund within the Interkadra project UDA-POKL-04.01.01-00-014/10-00 and Eu-Bridge 7th FR EU project (Grant Agreement No. 287658).

**References**

1. Koehn, P., Hoang, H., Birch, A., Callison-Burch, C., Federico, M., Bertoldi, N., Cowan, B., Shen, W., Moran, C., Zens, R., *et al*: **Open Source Toolkit for Statistical Machine Translation**. In: *ACL 2007: 2007*; 2007: 177-180.

2. Costa-jussà, M.R., and FMaS, J.: **Machine Translation in Medicine. A quality analysis of statistical machine translation in the medical domain** In: *Conference on Advanced Research in Scientific Areas (ARSA-2012): 2012*.
3. Kirchhoff, K., Turner, A.M., Axelrod, A., and Saavedra, F: **Application of statistical machine translation to public health information: a feasibility study**. *J Am Med Inform Assoc* 2011, **18**(4):473-478.
4. Dušek, O., Hajič, J., Hlaváčová, J., Novák, M., Pecina, P., Rosa, R., Tamchyna, A., Urešová, Z., and Zeman, D.: **Machine Translation of Medical Texts in the Khresmoi Project**. In: *Ninth Workshop on Statistical Machine Translation: 2014; Baltimore, MD, USA* Association for Computational Linguistics; 2014: 221-228.
5. Pletneva, N., and Vargas, A.: **Requirements for the general public health search**. In*.*, vol. Public Technical Report; 2011.
6. Goeuriot, L., Jones, G., Kelly, L., Kriewel, S., and Pecina, P.: **Report on and prototype of the translation support**. In*.*: Khresmoi Project; 2012.
7. Gschwandtner, M., Kritz, M., and Boyer, C.: **Requirements of the health professional search**. In*.*: Khresmoi Project 2011.
8. Worldwide H: **Medical Phrases and Terms Translation Demo**. In*.*; 1998 - 2015.
9. Karliner, L.S., Jacobs, E.A,, Chen, A.H., and Mutha, S.: **Do professional interpreters improve clinical care for patients with limited English proficiency? A systematic review of the literature**. *Health services research* 2007, **42**(2):727-754.
10. Schenker, Y., Perez-Stable, E.J., Nickleach, D., and Karliner, L.S.: **Patterns of interpreter use for hospitalized patients with limited English proficiency**. *Journal of general internal medicine* 2011, **26**(7):712-717.
11. Randhawa, G., Ferreyra, M., Ahmed, R., Ezzat, O., and Pottie, K.: **Using machine translation in clinical practice**. *Canadian family physician Medecin de famille canadien* 2013, **59**(4):382-383.
12. Deschenes, S.: **5 benefits of healthcare translation technology**. *Healthcare Finance News* 2012.
13. Zadon, C.: **Man Vs Machine: The Benefits of Medical Translation Services** In*.*: EzineArticles.com; 2013.
14. Tiedemann, J. (ed.): **News from OPUS - A Collection of Multilingual Parallel Corpora with Tools and Interfaces**. Amsterdam/Philadelphia: John Benjamins; 2009.
15. **Tokenization** [https://en.wikipedia.org/wiki/Tokenization_(lexical_analysis)]
16. Koehn, P., Hoang, H., Birch, A., Callison-Burch, C., Federico, M., Bertoldi, N., Cowan, B., Shen, W., Moran, C., Zens, R., *et al*: **Moses: Open Source Toolkit for Statistical Machine Translation**. In: *Annual Meeting of the Association for Computational Linguistics (ACL): 2007; Prague, Czech Republic*; 2007: 177–180.
17. Radziszewski, .A: **A tiered CRF tagger for Polish**. In: *Intelligent Tools for Building a Scientific Information Platform: Advanced Architectures and Solutions, editors: Membenik R, Skonieczny L, Rybiński H, Kryszkiewicz M, Niezgódka M, Springer Verlag, 2013*. edn. Edited by Membenik R. SL, Rybiński H., Kryszkiewicz M., Niezgódka M.: Springer Verlag; 2013.
18. **KantanMT - a sophisticated and powerful Machine Translation solution in an easy-to-use package** [http://www.kantanmt.com]

19. Lavie, A.: **Evaluating the Output of Machine Translation Systems**. In: *The Ninth Conference of the Association for Machine Translation in the Americas AMTA 2010* Denver, Colorado; 2010.
20. Axelrod, A.: **Factored Language Models for Statistical Machine Translation**: University of Edinburgh; 2006.
21. Koehn, P.: **What is a Better Translation? Reflections on Six Years of Running Evaluation Campaigns** *Tralogy* 2014(Session 5 - Quality in Translation / La qualité en traduction, Tralogy I).
22. Papineni, K., Rouskos S, Ward T, Zhu WJ: **BLEU: a Method for Automatic Evaluation of Machine Translation**. In: *40th Annual Meeting of the Assoc for Computational Linguistics: 2002; Philadelphia*; 2002: 311-318.
23. Doddington, G.: **Automatic Evaluation of Machine Translation Quality Using N-gram Co-Occurrence Statistics**. In: *Second International Conference on Human Language Technology (HLT) Research: 2002*; 2002: 138-145.
24. Banerjee, S., and Lavie, A.: **METEOR: An Automatic Metric for MT Evaluation with Improved Correlation with Human Judgments**. In: *Proc of ACL Workshop on Intrinsic and Extrinsic Evaluation Measures for Machine Translation and/or Summarization: 2005; Ann Arbor, USA*; 2005: 65-72.
25. Snover, M., Dorr, B., Schwartz, R., Micciulla, L., and Makhoul, J.: **A Study of Translation Edit Rate with Targeted Human Annotation**. In: *7th Conference of the Assoc for Machine Translation in the Americas, Cambridge, August 2006: 2006; Cambridge*; 2006.
26. Stolcke, A: **SRILM – An Extensible Language Modeling Toolkit**. In: *INTERSPEECH 2002: 2002*; 2002.
27. Koehn, P., Axelrod, A., Birch Mayne, A., Callison-Burch, C., Osborne, M., and Talbot, D.: **Edinburgh System Description for the 2005 IWSLT Speech Translation Evaluation**. In: *IWSLT: 2005*; 2005: 68-75.
28. Heafield, K.: **KenLM: Faster and smaller language model queries**. In: *Sixth Workshop on Statistical Machine Translation, Association for Computational Linguistics, 2011: 2011*: Association for Computational Linguistics; 2011.
29. Turner, A. M., Brownstein, M. K., Cole, K., Karasz, H., and Kirchhoff, K.: **Modeling workflow to design machine translation applications for public health practice**. *Journal of biomedical informatics* 2015, **53**:136-146.
30. Costa-jussà, M., and Fonollosa, J.: **Latest trends in hybrid machine translation and its applications**. *Computer Speech & Language* 2015, **32**(1):3-10.
31. Gao, Q., and Vogel, S.: **Parallel Implementations of Word Alignment Tool**. In: *ACL 2008 Software Engineering, Testing, and Quality Assurance Workshop: 2008*; 2008.
32. Durrani, N., Schmid, H., Fraser, A., Sajjad, H., and Farkas, R.: **Munich-Edinburgh-Stuttgart Submissions of OSM Systems at WMT13**. In: *ACL 2013 Eight Workshop on Statistical Machine Translation: 2013*; 2013.
33. Koehn, P., and Hoang, H.: **Factored Translation Models**. In: *Joint Conference on Empirical Methods in Natural Language Processing and Computational Natural Language Learning: 2007; Prague*; 2007: 868–876.
34. Bikel, D.: **Intricacies of Collins' Parsing Model**. In: *Association for Computational Linguistics: 2004*; 2004.

35. Dyer, C., Chahuneau V, Smith N: **A Simple, Fast and Effective Reparametrization of IMB Model 2**. In: *NAACL, 2013: 2013*; 2013.
36. Bojar, O., Buck, C., Callison-Burch, C., Federmann, C., Haddow, B., Koehn, P., Monz, C., Post, M., Soricut, R., and Specia, L.: **Findings of the 2013 Workshop on Statistical Machine Translation**. In: *Eight Workshop on Statistical Machine Translation: 2013; Sofia, Bulgaria*; 2013.
37. Hasan, A., Islam, S., and Rahman, M.: **A Comparative Study of Witten Bell and Kneser-Ney Smoothing Methods for Statistical Machine Translation**. *JU Journal of Information Technology (JIT)* 2012, **1**(1):1-6.
38. Wolk, K., and Marasek, K.: **Polish – English Speech Statistical Machine Translation Systems for the IWSLT 2013**. In: *10th International Workshop on Spoken Language Translation: 2013; Heidelberg, Germany*; 2013.
39. Tiedemann, J.: **Parallel Data, Tools and Interfaces in OPUS**. In: *8th International Conference on Language Resources and Evaluation (LREC 2012): 2012*; 2012.
40. Brants, T., and Franz A.: **Web 1T 5-gram version 1**. Philadelphia: Linguistic Data Consortium; 2006.